\documentclass{article}

\usepackage{arxiv}

\usepackage[utf8]{inputenc} 
\usepackage[T1]{fontenc}    
\usepackage{hyperref}       
\usepackage{url}            
\usepackage{booktabs}       
\usepackage{amsfonts}       
\usepackage{nicefrac}       
\usepackage{microtype}      
\usepackage{lipsum}
\usepackage{graphicx}

\usepackage{siunitx}

\usepackage{mathtools}

\usepackage{url}
\usepackage{tabularx}
\usepackage{multirow}
\usepackage{booktabs}
\usepackage{etoolbox}
\newrobustcmd{\B}{\bfseries}
\usepackage{color}
\usepackage[utf8]{inputenc}

\graphicspath{ {./images/} }

\title{General Demographic Pre-trained Models for Enhancing Predictive Performance Across Diseases and Population}

\author{
 Li-Chin Chen \\
  Data Analytics and Digital Transformation Research Center\\
  National Taiwan University\\
  Taipei, Taiwan\\
  \texttt{lichinc@ntu.edu.tw} \\
   \And
 Ji-Tian Sheu \\
  Department of Health Care Management\\
  Chang Gung University\\
  Taoyuan, Taiwan\\
  \texttt{jtsheu@mail.cgu.edu.tw} \\
  \And
 Yuh-Jue Chuang \\
  Department of Health Care Management\\
  Chang Gung University\\
  Taoyuan, Taiwan\\
  \texttt{chuangyj@mail.cgu.edu.tw} \\
}

\begin{document}
\maketitle
\begin{abstract}
\textbf{Background:} Foundation models for healthcare require balancing robust generalization across heterogeneous clinical populations and disease settings with the architectural simplicity needed for deployment. We present a pre-trained model focused on demographic attributes that enhances feature utility across medical domains in a plug-and-play fashion. \textbf{Methods:} We introduce the General Demographic Pre-trained (GDP) model, designed to extract intrinsic representations of patient status based on age and sex, the two most ubiquitous clinical features. The composition of GDP was optimized by investigating various encoding methods and visit-reordering schemes. The model was pre-trained and transferability was validated by embedding the learned representations into diverse disease and geographic cohorts characterized by distinct demographic profiles. The optimal model configuration was subsequently validated against top-performing tabular foundation models (TabPFN, TabICL, and TabFM). \textbf{Results:} Our findings demonstrate that concatenating GDP-derived embeddings with raw residual features consistently enhances predictive performance across classification tasks while elevating the relative importance of demographic attributes. The embedding transformation provides superior representational separability compared to the original data distribution, yielding competitive discrimination performance across metrics against all three general-purpose foundation models and tree-based algorithm. \textbf{Conclusions:} GDP has successfully served the purpose of a foundation model, which produce enriched representations that amplify the predictive insight of these features beyond their raw form. The generated embeddings can be directly concatenated with residual features, serving as an enhancement layer that maintains full compatibility with standard tabular classifiers. This framework offers a flexible, plug-and-play pathway to elevate downstream predictive performance while preserving computational simplicity and seamless integration with established machine learning workflows.
\end{abstract}


\section{Introduction}
The central challenge in healthcare foundation modeling lies in achieving robustness across heterogeneous clinical settings, diverse disease spectra, and varied demographic populations. Current research is commonly categorized according to data modalities, developed within homogeneous modalities such as natural language, vision, or audio. However, extending this paradigm to the healthcare domain reveals fundamental mismatches. First, modality-based classifications are insufficient for healthcare applications. A substantial proportion of clinically relevant information in Electronic Medical Records (EHR) are represented in structured, tabular form, including diagnoses, procedures, medications, and laboratory measurements, are frequently encoded using standardized terminologies such as ICD, LOINC, and SNOMED \cite{duan2024deep, li2022hiBEHRT}, these entities differ markedly in their semantic structure, temporal behavior, and clinical interpretation. Subsuming all such data under the broad category of “tabular” constitutes an overly coarse abstraction \cite{wornow2023shaky}. Second, the inherent heterogeneity of tabular data presents well-recognized challenges for the construction of foundation models. Unlike images, which exhibit strong spatial locality, or text, which follows sequential and syntactic regularities, tabular data integrate dense numerical variables with sparse categorical features, each reflecting distinct aspects of patient status \cite{harutyunyan2019multitask,TS2vec,GoolgleTimeFM_2023,borisov2022deep}. These variables are collected at disparate temporal resolutions, measured on heterogeneous scales, and often exhibit irregular sampling, missingness, and weak or nonstationary inter-feature dependencies \cite{borisov2022deep,somvanshi2024survey,HCNN2017,CNN_multifusion2016}. The absence of consistent spatial or semantic priors further complicates the extraction of robust and transferable representations from tabular data.

There is a need for modeling approaches that enable plug-and-play deployment in cross medical settings, diseases, and populations to boost the development and adoption of medical AIs in practice. Meanwhile, demographic attributes, such as age, sex, or educational background, constitute foundational components of patient data in heterogeneous clinical settings. These attributes are highly standardized, widely available, and intrinsically informative. Therefore, we propose developing pre-trained models grounded in EHR components, in this instance, the demographic attributes, and demonstrate its ability to improve predictive performance across diverse clinical environments, disease contexts, and patient populations, while maintaining simplicity in deployment.

\subsection{Common Processing and Encoding of EHR data and Demographic Attributes in Machine Learning}

Demographic attributes are fundamental and readily available patient characteristics. Age functions as a critical determinant, providing signals of biological vulnerability, diagnostic framing, therapeutic constraints, risk stratification, and eligibility for age-specific screening \cite{jackson2004burden,mandelblatt2021applying,american2012patient}. Similarly, gender or sex constitutes a fundamental axis along which patients may exhibit differing physiological responses to a wide range of diseases \cite{oertelt2009gender,courchesne2023gender,moores2023sex}. Despite their ubiquity, the demographic attributes were genearally seen as an auxiliary part of the EHR data. For instance, Wornow et al. \cite{wornow2023ehrshot} and Hur et al. \cite{hur2023genhpf} both transformed all medical events into natural language descriptions and then tokenized them into embeddings via a language model encoder, whereas Wornow et al. converted medical codes into discretized value ranges (e.g. Weight/180-190) and Hur et al. expressed them as feature name and its value (e.g. vancomycin: 10.0). Meanwhile, although Yang et al. \cite{yang2023transformehr} did treat different entities in the EHR differently and received different embeddings (e.g. visit embeddings, temporal embeddings, and code/demographic embeddings), the embedding were eventually summed up to became the model input. Some studies encode age as sinusoidal sequences via Fourier-based transformations and integrate them with other concept embeddings \cite{fallahpour2024ehrmamba,kazemi2019time2vec}, whereas others omit demographic attributes \cite{2021MEDBERT}. Although the aforementioned studies achieved strong performance on specific tasks, their models were not designed to be deployed with established machine learning workflows across heterogeneous populations and disease settings.

\subsection{Transferability across clinical settings, diseases, and demographic populations}

The transferability of foundation models refers to their capacity to be trained on large-scale datasets, to extract high-level abstractions from raw inputs, and to leverage the resulting embeddings to enhance downstream predictive performance \cite{borisov2022deep,MyGLP,yoon2020vime,fallahpour2024ehrmamba,rasmy2021medBERT,yuanyuan2025scoping}. Prior work has demonstrated cross-disease transferability, such as training retinal image–specific models that can be repurposed for the diagnosis of both ocular diseases and systemic disorders \cite{zhou2023foundation}, as well as learning general-purpose histopathology slide representations that enable pathology report generation and generalize to resource-limited clinical scenarios, including rare disease retrieval and cancer prognosis \cite{ding2025multimodal}. However, recent studies have emphasized that the transferability of existing systems still requires careful adaptation rather than plug-and-play deployment \cite{ramanathan2025abstract}. This limitation primarily arises from substantial heterogeneity in data distributions across populations, geographic regions, medical centers, data collection devices, and clinical protocols, which can lead to distributional mismatch between foundation models and downstream tasks, thereby hindering effective generalization \cite{he2024foundation}. To bridge the gap of heterogeneity in data, many prior studies have focused on encoding EHR data and reorganizing patient visits into sequential representations \cite{2021MEDBERT, yang2023transformehr, hur2023genhpf,yuanyuan2025scoping,borisov2022deep}. 

Concurrently, a growing body of research aims to develop general-purpose tabular models that consistently outperform tree-based algorithms across diverse domain contexts \cite{erickson2026tabarena}. The primary focus within this paradigm leverages in-context learning to eliminate the need for task-specific hyperparameter tuning, as exemplified by architectures such as TabPFN  \cite{hollmann2025accurate}, TabICL \cite{qutabular}, and TabFM \cite{googleresearch2026tabfm_blog}. This study aims to segment EHR components, and develop a General Demographic Pre-trained (GDP) model, which enhanced the predictive utility of demographic attributes across multiple medical prediction tasks and heterogeneous clinical settings, and also maintained simplicity and seamless compatibility with established machine learning pipelines.

\subsection{Methods}

\begin{figure*}
\centerline{\includegraphics[width=0.8\columnwidth]{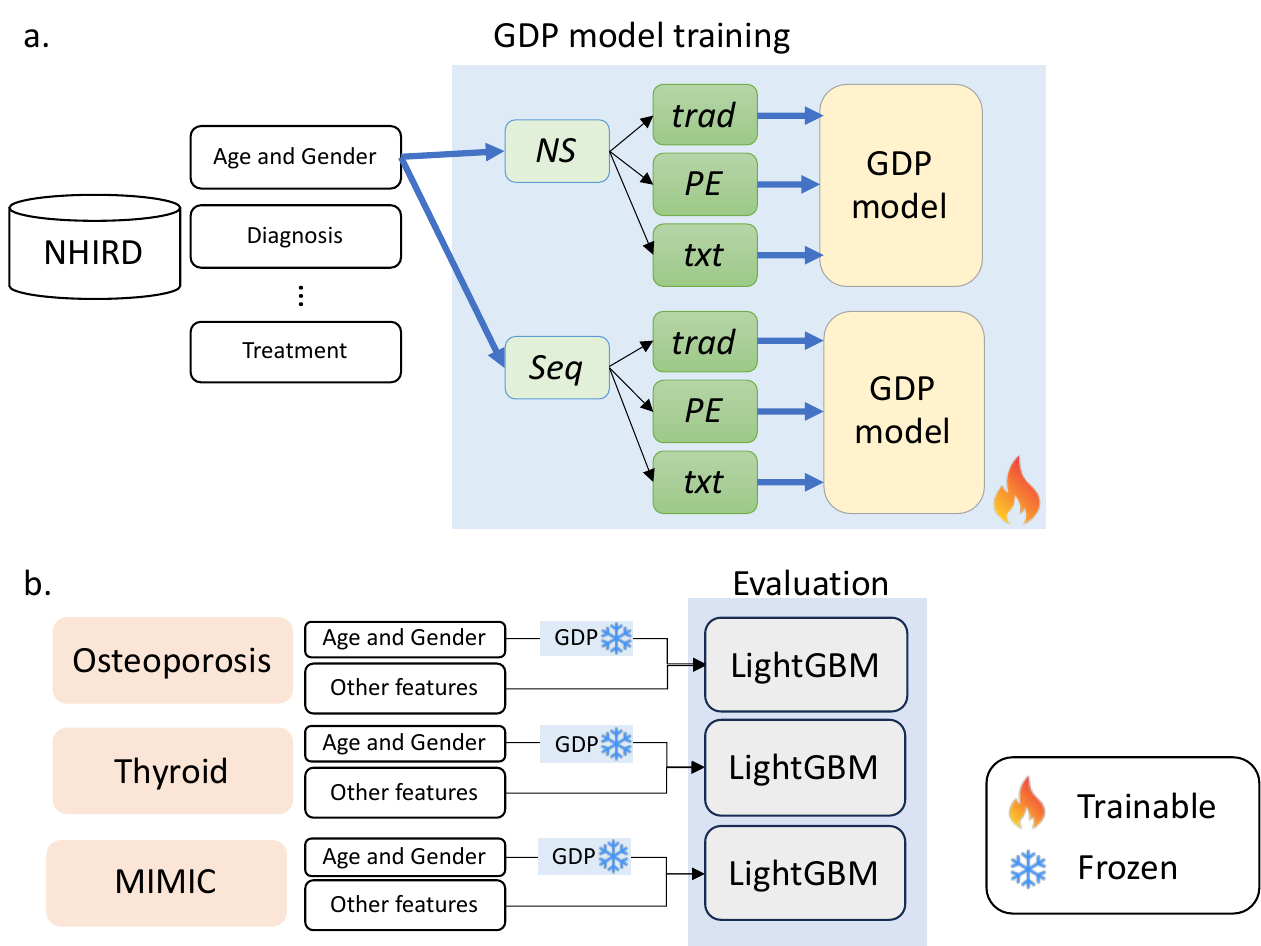}}
\caption{General demographical pre-trained model training and validation flow. GDP: General Demographic Pre-trained; NHIRD: National Health Insurance Research Database of Taiwan; $NS$: non-sequential ordering; $Seq$: Sequential ordering; $\emph{trad}$: traditional encoding; $\emph{PE}$: positional encoding; $\emph{txt}$: text-based semantic encoding.}
\label{training_flow}
\end{figure*}

The intention of this work was to train a model that captured the abstract correlation between the demographic attributes and the disease severity, and then transfer it to diverse scenario and populations. The overall workflow is illustrated in Figure~\ref{training_flow}. GDP model was initially pre-trained on age and gender information to predict the Charlson Comorbidity Index (CCI) \cite{charlson1987new} at each patient visit. The CCI is a widely adopted measure for assessing patient mortality risk and disease severity based on comorbid conditions, and thus serves as a robust proxy for generating embeddings that capture clinically meaningful health status representations. Demographic variables (age and gender) and the three primary diagnoses were recorded for each patient encounter to compute the corresponding CCI score. Longitudinal patient data were ordered chronologically by age at the time of the visit and structured into a 120-month (10-year) window. Records exceeding 120 months were truncated, whereas those under 120 months were zero-padded to maintain uniform input dimensions. Using the derived CCI scores as self-generating prediction targets, we trained the model in a self-supervised learning framework without manual labeling \cite{he2024foundation, bommasani2021opportunities, moor2023foundation,2024ehrshot_followup,borisov2022deep}. The pre-training process in shown as in Figure~\ref{training_flow}a. The transferability of the model was evaluated through applied the demographic attributes to four medical predictive tasks that involve heterogeneous diseases, population, and clinical settings (Figure~\ref{training_flow}b). Let $\mathcal{D} = \{(\mathbf{x}_i, y_i)\}_{i=1}^N$ be the pre-training dataset of $N$ patients, $\mathbf{x} \in \mathcal{X}$ represent the input vector of demographic attributes (e.g., age and gender) for a patient visit, and $y \in \mathcal{Y}$ represent the calculated CCI. The objective is to train a model $f_\theta: \mathcal{X} \rightarrow \mathcal{Y}$, parameterized by $\theta$, to capture the abstract correlation between demographics and disease severity. The optimization problem to learn the initial parameters $\theta_{\text{pre}}$ can be written as:$$\theta_{\text{pre}} = \arg\min_{\theta} \frac{1}{N} \sum_{i=1}^N \mathcal{L}\big(f_\theta(\mathbf{x}_i), y_i\big)$$, where $f_\theta(\mathbf{x}_i)$ is the GDP model's prediction, and $\mathcal{L}(\cdot)$ is the loss function. To evaluate the relative impact of demographic encoding and temporal sequence ordering on model training, this study systematically benchmarks three distinct encoding approaches alongside two visit-reordering schemes to identify the optimal architectural configuration. Once the optimal setup for GDP is established, it is evaluated against state-of-the-art general-purpose foundation models, namely TabPFN, TabICL, and TabFM. These approaches represent top-performing baselines that regularly anchor the upper tier of the TabArena leaderboard \cite{erickson2026tabarena}.

\subsection{Data Encoding and Sequential Ordering Strategies}
Drawing upon established methodologies from previous research, this study investigates data encoding and ordering schemes that are widely recognized for enhancing machine learning performance\cite{wornow2023ehrshot,hur2023genhpf,yang2023transformehr,vaswani2017attention}. The first encoding approach is the traditional encoding ($\emph{trad}$), gender ($g$) was represented using one-hot encoding, where $g \in \{0, 1\}$, and age ($x$) was transformed with a natural logarithm $x_e = \log(1 + x)$, where $x \in \mathbb{R}^n$ and $e$ denotes the resulting embedding vector. The second approach is the positional encoding ($\emph{PE}$), embedded age information using a sinusoidal positional encoding scheme and further differentiated it by adding zeros or ones for each gender, which is expressed as:

\begin{equation}
PE_{(pos, 2i)} = \sin\left(\frac{pos}{10000^{2i/d_{model}}}\right) + g_{i} \\
\end{equation}

\begin{equation}
PE_{(pos, 2i+1)} = \cos\left(\frac{pos}{10000^{2i/d_{model}}}\right) + g_{i}
\end{equation}, where $pos$ is the position index, $i$ is the dimension index, $g_i \in \{0,1\}$, and $g \in \mathbb{R}^{d_{\text{model}}}$. Finally, in the text-based semantic encoding ($\emph{txt}$), demographic information was first expressed as short descriptive text strings (e.g., Male, 75 years old) and subsequently converted into embeddings using the encoder of an open language model:

\begin{equation}
E_{w_i} = f_{encoder}([w_1, \dots, w_i]) 
\end{equation}, $w_i$ denotes the $i^\text{th}$ token, $E_{w_i} \in \mathbb{R}^d$ is its embedding vector, and $d$ represents the embedding dimension. The encoder used was the all-MiniLM-L6-v2 model, accessed via the pymilvus Python package \cite{wang2021milvus}.

The two evaluated ordering schemes were the non-sequential ($\emph{NS}$) and the sequential ($\emph{Seq}$) ordering schemes. In $\emph{NS}$, patient visits were arranged randomly, whereas in $\emph{Seq}$, visits were sorted by age, allowing multiple rows per year. Each patient was assigned exclusively to either the training or testing set to avoid data leakage.

\subsection{Model Architecture}

The model architecture was tailored to the chosen ordering scheme. For the $\emph{NS}$ configuration, the model consisted of a linear layer followed by an attention mechanism and two additional linear layers, each with ReLU activations in between. For the $\emph{Seq}$ configuration, the model was composed of a single-layer long short-term memory (LSTM) network followed by two linear layers separated by a ReLU activation. Each ordering scheme ($\emph{NS}$ and $\emph{Seq}$) was combined with an encoding strategy ($\emph{trad}$, $\emph{PE}$, and $\emph{txt}$), yielding six candidate configurations for GDP. 

GDP pre-training was conducted using a representative subset of two million clinical claims and registry records from the National Health Insurance Research Database (NHIRD) of Taiwan, covering the observation window from January 1, 2002, to December 31, 2011. Records with missing birth date, gender, or diagnosis information were excluded, and diagnosis codes were used solely for generating the CCI labels, not as model inputs. A 32-dimensional latent representation is extracted from the penultimate layer and subsequently concatenated with the downstream feature vectors.

\subsection{Transferability Assessment}

Following the pre-training phase, model transferability was evaluated by incorporating the learned embeddings into three distinct binary classification tasks: (1) osteoporosis prediction utilizing a public dataset from the United States obtained via Kaggle\footnote{\url{https://www.kaggle.com/code/supriyoain/osteoporosis-xgbclassifier-91-5-accuracy/input}}; (2) thyroid disease classification using an OpenML dataset of Australian origin\footnote{\url{https://www.openml.org/search?type=data&sort=runs&status=active&id=38}}; and (3) 30-day post-discharge mortality prediction based on US patient data from the MIMIC-III database \cite{mimiciii}, spanning diagnostic categories across cardiovascular (50.41\%), neurological (12.01\%), respiratory (9.98\%), gastrointestinal (9.08\%), infectious (6.71\%), renal/genitourinary (4.51\%), trauma (3.96\%), endocrine/metabolic (1.86\%), and hematologic/oncologic (1.48\%) diseases. Collectively, these datasets encompass diverse disease domains and exhibit substantial heterogeneity regarding demographics and patient populations. 

For each of the three classification tasks, demographic attributes were replaced with GDP-derived embeddings and concatenated with the remaining features. All the remaining features were processed using standard preprocessing procedures: categorical variables were encoded via one-hot encoding, and numerical variables were transformed using $x_e = \log(1 + x)$. In each case, LightGBM served as the predictive model for downstream tasks owing to its strong performance on tabular data \cite{ke2017lightgbm}. The comparison baseline was established using the raw datasets, and classified with LightGBM. Hyperparameter settings of LightGBM and  data aggregation details for MIMIC and pneumonia datasets were summarized in Appendix. 

Performance was measured using the area under the receiver operating characteristic curve (AUROC). Results were averaged over 50 bootstrap samples, and statistical significance was determined using independent $\emph{t-tests}$ with a significance threshold of $p < 0.05$. To visualize the changes of the learned embeddings, age and gender vectors produced by the GDP model were projected into two dimensions and demonstrated using t-distributed Stochastic Neighbor Embedding (t-SNE). Further, changes in the feature importance of demographic attributes were quantified by analyzing LightGBM’s information gain before and after incorporating the GDP embeddings. 

Upon determining the optimal configuration for GDP, we benchmarked it against TabPFN (version 6.3.0), TabICL (version 2.1.1), and TabFM (version 1.0.1). Due to the fact that these baseline models generate final point predictions directly, we also evaluated all models using a 50-sample bootstrap averaging procedure, followed by independent $\emph{t-tests}$ to assess statistical significance. Performance was measured using the AUROC and the Expected Calibration Error (ECE). ECE was additionally evaluated to assess how well each model's estimated probabilities aligned with true empirical likelihoods. This study was approved by the Research Ethics Committee at National Taiwan University (No. 202409HM027) and waived the requirement for informed patient consent for the data, which had already been de-identified before analysis.

\section{Results}
\subsection{Pre-trained and validated datasets}
The pre-training cohort comprised 130,000 patients with a total of 11,551,582 visit records. Among them, 44.28\% were male patients (\textit{n} = 57,561). Across longitudinal patient records, the median age at clinical visits ranged from a median of 35 years {[}13, 52{]} to 46 years {[}24, 63{]}, and the median CCI score ranged from 0 {[}0, 1{]} to 0 {[}0, 2{]}\footnote{The values are presented as median {[}first quartile (Q1), third quartile (Q3){]}}. Based on the age-sex interaction analysis, age is the dominant predictor of CCI accumulation, and this relationship is consistent across sexes. There are minimal to negligible age-sex interaction due to visual figures demonstrates parallel and identical trajectories in linear, age binning, and spline analysis, the trajectories for males and females remained parallel at every age stage without crossing, divergence, or separate. Further detailed interpretation and visualization figures are shown in Appendix. 

Table~\ref{table_demographic} summarizes the demographic distributions across the validation datasets. The patient populations differed across the three cohorts. Thyroid disease and MIMIC datasets were skewed toward older patients, whereas the osteoporosis cohort comprised younger individuals. Gender distributions also varied, with a higher proportion of female patients in the thyroid dataset and a higher proportion of male patients in the MIMIC dataset. Outcome labels were generally balanced across all cohorts.

Table~\ref{table_demographic} further presents the importance of age and gender, which was quantified by summing its information gain contributions across all splits in baseline LightGBM and normalizing this value by the total information gain of all features, yielding a percentage that represents the feature’s relative weight. A ranking was further assigned to indicate the relative importance of features, ordered in descending order. Results indicate that age was a highly influential feature in all three datasets, ranking within the top three influential feature. Gender consistently exhibited limited predictive value. Although ranked fifth in the thyroid disease dataset, gender accounted for only 2.00\% of the total weight, and its contribution was even smaller in the other two datasets.

\begin{table*}
\caption{Demographic distribution of validation dataset}
\label{table_demographic}
\setlength{\tabcolsep}{3pt}
\centering
\begin{tabular}{p{66pt}p{30pt}p{55pt}p{55pt}p{66pt}p{70pt}p{80pt}}
\toprule
             & \textit{n}    & Age, {[}\text{Q1, Q3}{]} & Male, \textit{n} (\%) & Outcomes, \textit{n} (\%)  & Age Importance, (\%) (Rank) & Gender Importance, (\%) (Rank) \\
\toprule
Osteoporosis & 1,958 & 32 {[}21, 53{]} & 992 (50.66) & 979 (50.00) & 84.93 (1/13) & 1.11 (10/13) \\
Thyroid      & 450   & 60 {[}46, 72{]} & 169 (37.56) & 225 (50.00) & 29.58 (2/22) & 2.00 (5/22) \\
MIMIC        & 1,033 & 66 {[}56, 76{]} & 616 (59.63) & 463 (44.82) & 14.60 (3/8) & 1.88 (8/8) \\
\bottomrule
\multicolumn{7}{p{465pt}}{\textit{n}: number of samples; Q1: first quartile; Q3: third quartile; The percentage of importance was computed by dividing the information gain contributions across all splits in LightGBM of each feature by the total gain of all features, and the result was expressed as a percentage; Rank: indicate the relative importance of features, ordered in descending order.}
\end{tabular}
\end{table*}

\subsection{Transferability Results}

Tables~\ref{table_osteoporosis} to \ref{table_MIMIC} present the predictive performance across the three datasets. In all three datasets, the $\emph{Seq}$ ordering achieved significant performance gains ($p < $0.05), demonstrating robustness irrespective of the encoding scheme utilized. $\emph{Seq}$ attained near-perfect discrimination (mean AUROC = 1.000) when rounded to three decimal places. When comparing the three encoding strategies, results varied between diseases. In the osteoporosis dataset, the AUROC values did not reach significantly differences among encodings ($\emph{Seq-trad}$ and $\emph{Seq-PE}$, $p$ = 0.906; $\emph{Seq-trad}$ and $\emph{Seq-txt}$, $p$ = 0.961; and $\emph{Seq-PE}$ and $\emph{Seq-txt}$, $p$ = 0.819), whereas in the thyroid and MIMIC dataset, the AUROC values differed significantly across all three encoding strategies ($p <$ 0.001), having $\emph{Seq-trad}$ outperformed $\emph{Seq-txt}$ ($p <$ 0.001), and $\emph{Seq-txt}$ outperformed $\emph{Seq-PE}$ ($p <$ 0.001).

\begin{table*}
\caption{Osteoporosis dataset prediction results}
\label{table_osteoporosis}
\setlength{\tabcolsep}{3pt}
\centering
\begin{tabular}{p{30pt}p{85pt}p{28pt}p{28pt}p{85pt}p{28pt}p{28pt}p{85pt}p{28pt}p{28pt}}
\toprule
& $\emph{trad}$ & $\emph{p-value}^{a}$ & $\emph{p-value}^{b}$ & $\emph{PE}$ & $\emph{p-value}^{a}$ & $\emph{p-value}^{b}$ & $\emph{txt}$  & $\emph{p-value}^{a}$ & $\emph{p-value}^{b}$ \\
\midrule
\multicolumn{9}{c}{AUROC (Mean ± 95\% CI)}\\
\midrule
Baseline & 0.921 {[} 0.918, 0.925 {]} & & & & \\
\textit{NS} & 0.917 {[} 0.913, 0.921 {]} & 0.113 & & 0.909 {[} 0.905, 0.913 {]} & $<$0.001* & & 0.852 {[} 0.847, 0.857 {]} & $<$0.001* & \\
\textit{Seq} & \textbf{1.000} {[} 1.000, 1.000 {]} & $<$0.001* & $<$0.001* & \textbf{1.000} {[} 1.000, 1.000 {]} & $<$0.001* & $<$0.001* & \textbf{1.000} {[} 1.000, 1.000 {]} & $<$0.001* & $<$0.001* \\
\bottomrule
\multicolumn{9}{p{465pt}}{The best-performing values are highlighted in bold. \textit{Seq} attained 1.000 when rounded to three decimal places. The AUROC comparisons between \textit{Seq-trad} and \textit{Seq-PE} (\textit{p} = 0.906), \textit{Seq-trad} and \textit{Seq-txt} (\textit{p} = 0.961), and \textit{Seq-PE} and \textit{Seq-txt} (\textit{p} = 0.819) indicate no statistically significant differences. $\textit{p-value}^{a}$: t-test results between baseline and \textit{NS}, and \textit{Seq}; $\textit{p-value}^{b}$: t-test results between \textit{NS} and \textit{Seq}.}
\end{tabular}
\end{table*}

\begin{table*}
\caption{Thyroid disease dataset prediction results}
\label{table_Thyroid}
\setlength{\tabcolsep}{3pt}
\centering
\begin{tabular}{p{30pt}p{85pt}p{32pt}p{32pt}p{85pt}p{28pt}p{28pt}p{85pt}p{28pt}p{28pt}}
\toprule
& $\emph{trad}$ & $\emph{p-value}^{a}$ & $\emph{p-value}^{b}$ & $\emph{PE}$ & $\emph{p-value}^{a}$ & $\emph{p-value}^{b}$ & $\emph{txt}$  & $\emph{p-value}^{a}$ & $\emph{p-value}^{b}$ \\
\midrule
\multicolumn{9}{c}{AUROC (Mean ± 95\% CI)}\\
\midrule
Baseline & 0.831 {[} 0.818, 0.843 {]} & & & & & &\\
\textit{NS} & 0.842 {[} 0.831, 0.854 {]} & 0.169 & & 0.816 {[} 0.804, 0.828 {]} & 0.093 & & 0.827 {[} 0.817, 0.838 {]} & 0.693 & \\
\textit{Seq} & \textbf{0.997} {[} 0.995, 0.999 {]} & $<$0.001* & $<$0.001* & \textbf{0.988} {[} 0.986, 0.990 {]} & $<$0.001* & 0.002* & \textbf{0.993} {[} 0.991, 0.994 {]} & $<$0.001* & $<$0.001* \\
\bottomrule
\multicolumn{9}{p{465pt}}{The best-performing values are highlighted in bold. The AUROC differences between \textit{Seq-trad} and \textit{Seq-PE} (\textit{p} $<$ 0.001*), \textit{Seq-trad} and \textit{Seq-txt} (\textit{p} $<$ 0.001*), and \textit{Seq-PE} and \textit{Seq-txt} (\textit{p} = 0.001*) are statistically significant. $\textit{p-value}^{a}$: t-test results between baseline and \textit{NS}, and \textit{Seq}; $\textit{p-value}^{b}$: t-test results between \textit{NS} and \textit{Seq}.}
\end{tabular}
\end{table*}

\begin{table*}
\caption{MIMIC dataset prediction results}
\label{table_MIMIC}
\setlength{\tabcolsep}{3pt}
\centering
\begin{tabular}{p{30pt}p{85pt}p{28pt}p{28pt}p{85pt}p{28pt}p{28pt}p{85pt}p{28pt}p{28pt}}
\toprule
& $\emph{trad}$ & $\emph{p-value}^{a}$ & $\emph{p-value}^{b}$ & $\emph{PE}$ & $\emph{p-value}^{a}$ & $\emph{p-value}^{b}$ & $\emph{txt}$  & $\emph{p-value}^{a}$ & $\emph{p-value}^{b}$ \\
\midrule
\multicolumn{9}{c}{AUROC (Mean ± 95\% CI)}\\
\midrule
Baseline & 0.610 {[} 0.592, 0.637 {]}  & & & & & \\
\textit{NS} & 0.617 {[} 0.600, 0.637 {]} & 0.303 & & 0.607 {[} 0.585, 0.631 {]} & 0.612 & & 0.606 {[} 0.583, 0.626{]} & 0.552\\
\textit{Seq} & \textbf{0.695} {[} 0.677, 0.719 {]} & $<$0.001* & $<$0.001* & \textbf{0.625} {[} 0.604, 0.643 {]} & 0.041* & 0.008*  & \textbf{0.671} {[} 0.659, 0.692 {]} & $<$0.001* & $<$0.001* \\
\bottomrule
\multicolumn{9}{p{470pt}}{The best-performing values are highlighted in bold. The AUROC differences between \textit{Seq-trad} and \textit{Seq-PE} (\textit{p} $<$ 0.001*), \textit{Seq-trad} and \textit{Seq-txt} (\textit{p} $<$ 0.001*), and \textit{Seq-PE} and \textit{Seq-txt} (\textit{p} $<$ 0.001*) are statistically significant. \textit{NS}: non-sequential approack; \textit{Seq}: sequential approach; $\textit{p-value}^{a}$: t-test results between baseline and \textit{NS}, and \textit{Seq}; $\textit{p-value}^{b}$: t-test results between \textit{NS} and \textit{Seq}.}
\end{tabular}
\end{table*}

\subsection{Representation Distribution Changes}
\begin{figure*}
\centerline{\includegraphics[width=0.8\columnwidth]{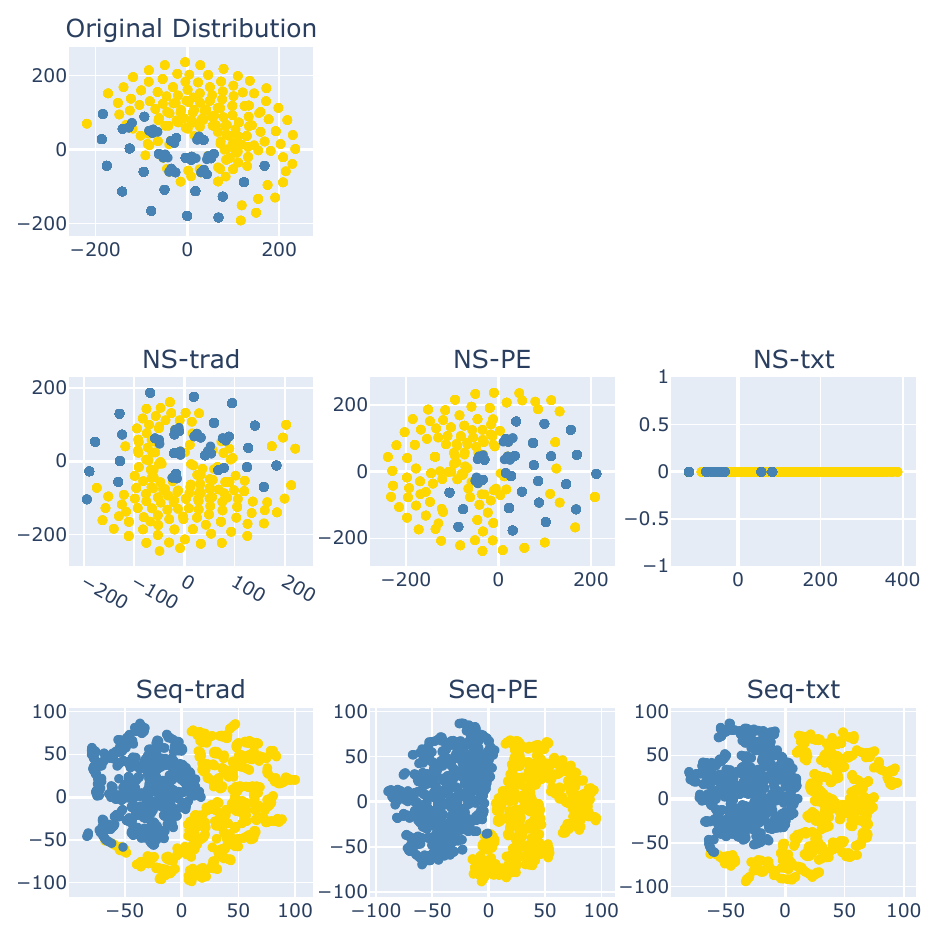}}
\caption{Representation distribution of osteoporosis dataset.}
\label{osteoporosis_embedding}
\end{figure*}

\begin{figure*}
\centerline{\includegraphics[width=0.8\columnwidth]{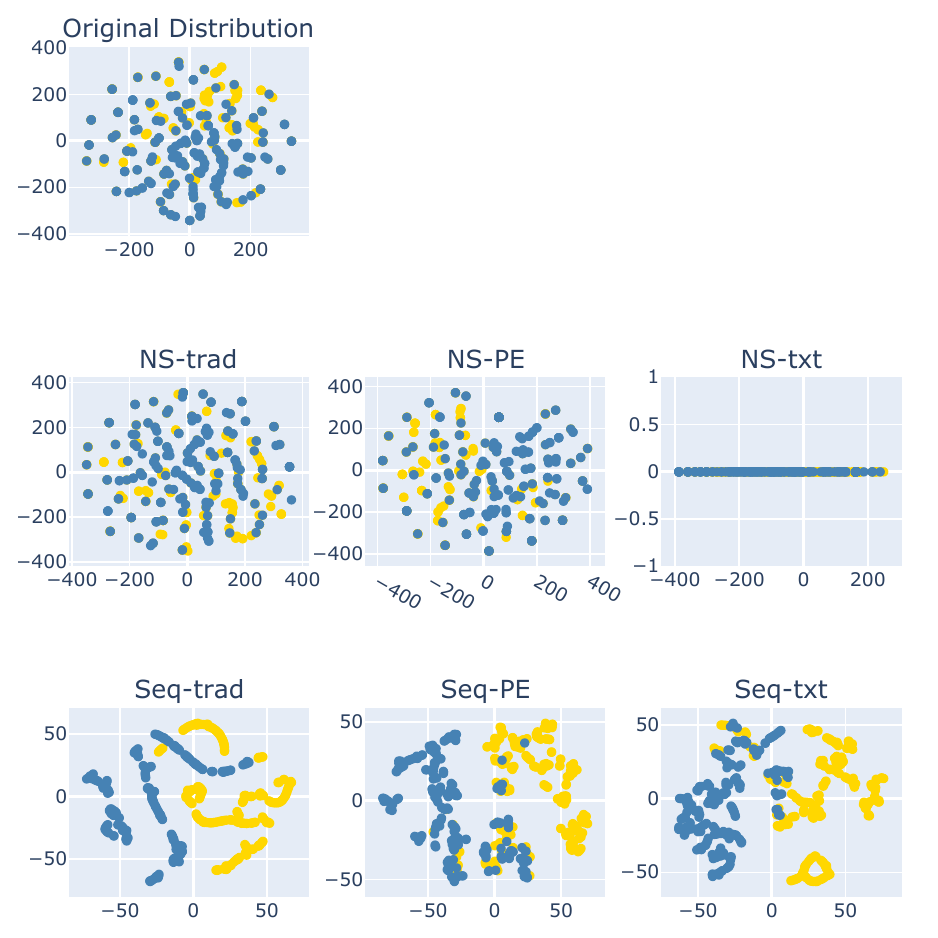}}
\caption{Representation distribution of thyroid dataset.}
\label{ThyroidDisease_embedding}
\end{figure*}

\begin{figure*}
\centerline{\includegraphics[width=0.8\columnwidth]{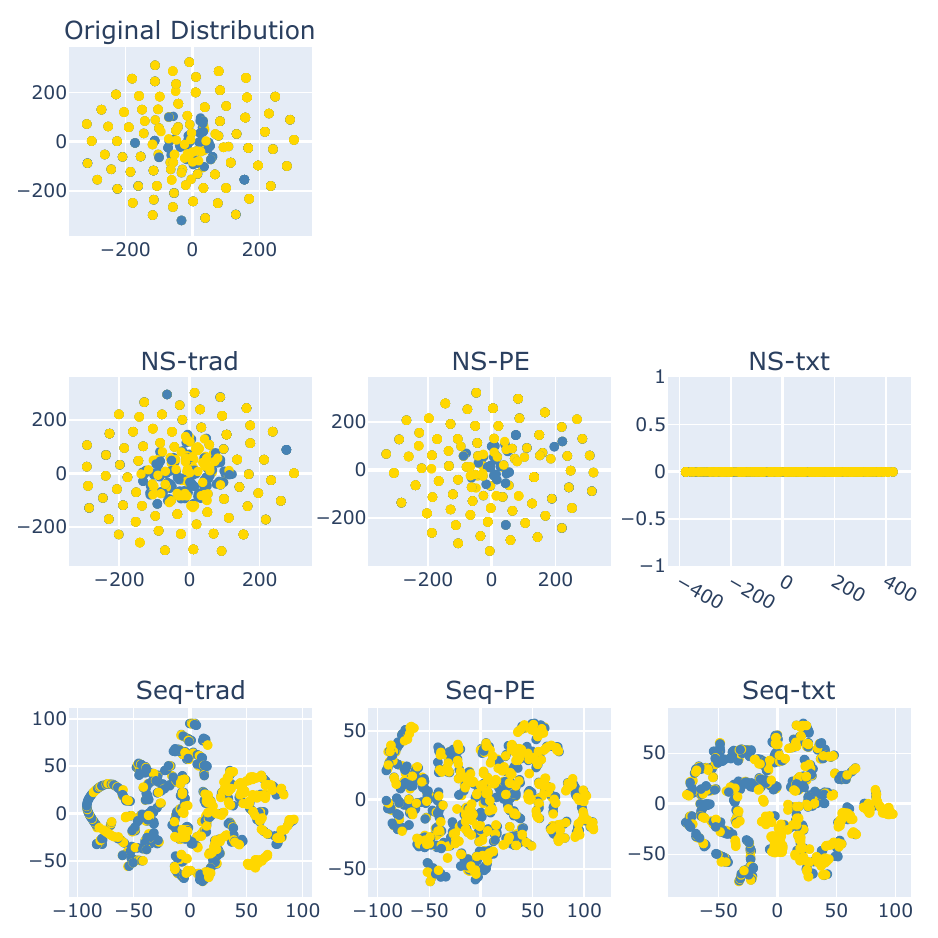}}
\caption{Representation distribution of MIMIC dataset.}
\label{MIMIC_embedding}
\end{figure*}

Figure~\ref{osteoporosis_embedding} to \ref{MIMIC_embedding} illustrate the learned representations under different approaches after dimensionality reduction with t-SNE. Generally, the $\emph{Seq}$ approach induced effective transformation in the representation space, compressing the variation of data into a narrower range and more clearly separating the two outcome labels. By contrast, the distribution generated by the $\emph{NS}$ approach either exhibited minimal deviation from the original distribution or yielded variations that failed to enhance classification prediction. For both \textit{NS-trad} and \textit{NS-PE}, the value ranges were maintained close alignment with the original, whereas in the \textit{NS-txt} approach, t-SNE tended to project the second dimension toward zero.

\subsection{Feature Importance Changes}

\begin{figure*}
\centerline{\includegraphics[width=1.0\columnwidth]{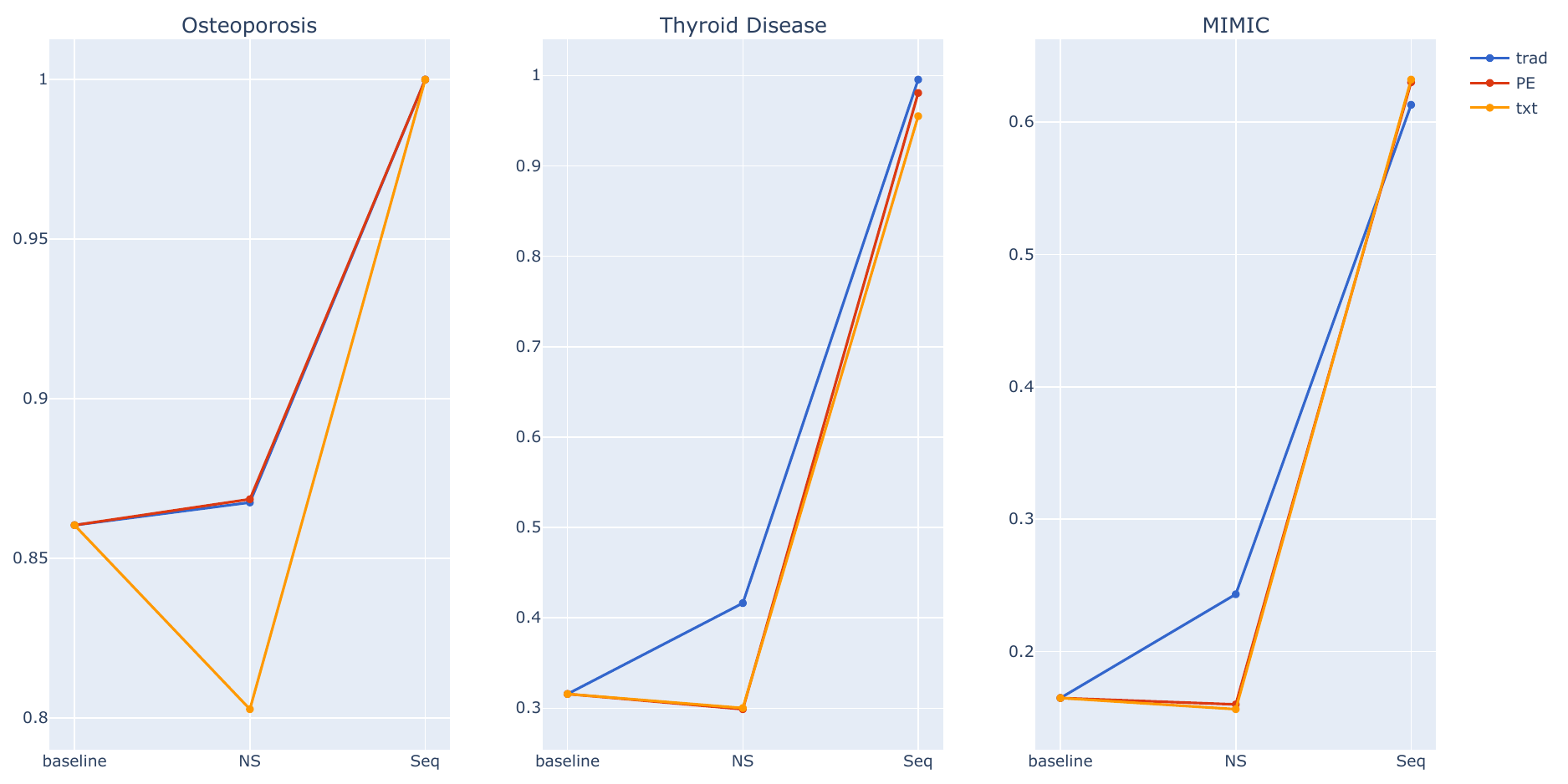}}
\caption{Comparison of age and gender information gain across approaches. Scores reflect the aggregate contribution of age and gender, divided by the overall information gain of all features.}
\label{feature_importance}
\end{figure*}

Figure~\ref{feature_importance} depicts the variation in information gain after applying GDP while validating with LightGBM. Across all datasets, the $\emph{Seq}$ approach consistently increased the relative importance of demographic attributes compared with the baseline. In contrast, the $\emph{NS}$ approach yielded less stable outcomes: in some cases, it enhanced the importance beyond baseline levels, while in others it diminished it. Across the three encoding schemes for $\emph{Seq}$, the aggregated feature importance were nearly identical.

\subsection{Comparison with Top-performing General-purpose Foundation Models}

\begin{figure*}
\centerline{\includegraphics[width=1.0\columnwidth]{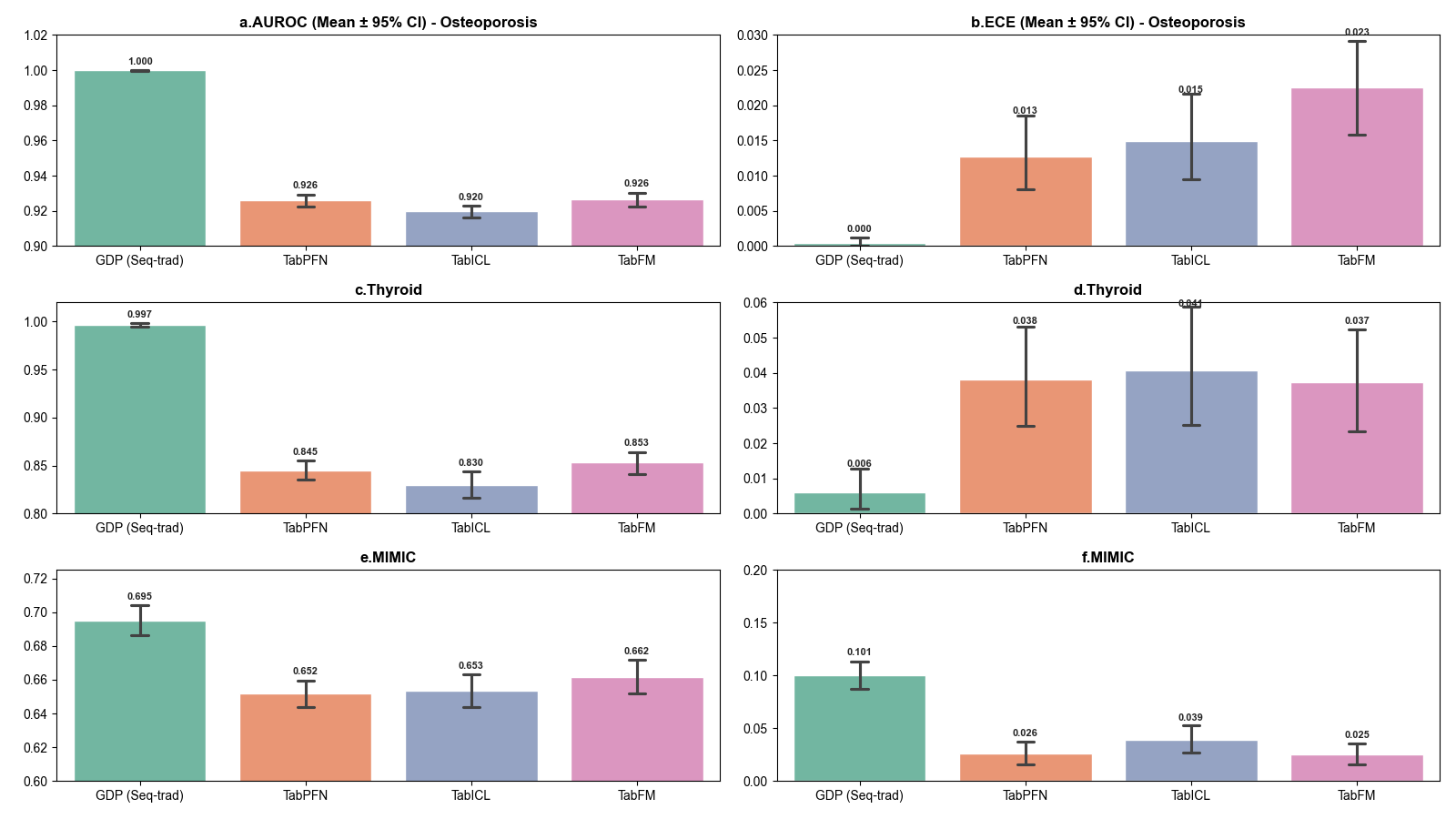}}
\caption{Comparative performance between the optimal GDP configuration and general-purpose foundation models. Across both discriminative performance and probability calibration, all pairwise performance differences between GDP and the foundation models reached statistical significance (\textit{p} $<$ 0.001).}
\label{external_model_compare}
\end{figure*}

Having identified $\emph{Seq-trad}$ as the best-performing configuration, it is further benchmarked against the general-purpose models, shown in Figure~\ref{external_model_compare}. Across all three datasets, GDP achieved superior discriminative performance compared to foundation models (TabPFN, TabICL, and TabFM), having mean AUROC statistical significantly higher (\textit{p} $<$ 0.001) than the foundation models. Regarding probability calibration, GDP demonstrated near-perfect alignment on the osteoporosis (ECE = 0.000, \textit{p} $<$ 0.001) and thyroid (ECE = 0.006, \textit{p} $<$ 0.001) cohorts. However, on the MIMIC dataset, while GDP retained the highest AUROC, it demonstrates higher ECE (ECE = 0.101, \textit{p} $<$ 0.001) relative to other models, suggesting a trade-off between discrimination and probability calibration under high data variability.

\section{Discussion}

While the versatility of general-purpose foundation models makes them highly attractive across diverse scenarios, model generalization often comes at the expense of domain-specific accuracy. Given that high precision is critical in healthcare, conversely, overly task-specific models introduce considerable development burdens and induce substantial domain gap, such as the challenges presented when directly applying the Segment Anything Model (SAM) \cite{SAM2023} to medicine \cite{noh2025narrative}. Further attempts to deploy general-purpose models to reduce domain-specific engineering costs have shown mixed results \cite{LLM_in_medicine_review2023ACM}. Rather than relying on generic paradigms, clinical machine learning demands foundation models engineered specifically to capture fundamental patient dynamics, that are invariant to disease, setting, or population, while remaining flexibility to integrate into existing healthcare applications. 

This study sought to explore the development of a foundation model tailored to demographic attributes of patients, with the expectation that such a model could provide generalizable enhancements to predictive performance across tasks, irrespective of disease type or population differences in a plug-and-play fashion. Our findings demonstrate that concatenating GDP-derived embeddings with the remaining features successfully enhances predictive performance across the classification tasks while simultaneously elevating the feature importance of demographic attributes. The GDP model enhanced representational separability compared with the original distribution, which translated into modest improvements in discrimination performance and increases in information gain during node splitting. Performance across metrics is competitive with all three general-purpose foundation models; however, slightly elevated calibration metrics on complex disease cohorts suggest an inherent trade-off between discriminative power and probability calibration when handling highly variable data.

Although accessing large-scale, publicly available datasets that offer comprehensive disease spectra alongside population diversity remains a critical challenge \cite{wornow2023shaky,2024ehrshot_followup, hur2023genhpf, moor2023foundation, he2024foundation}, this study utilizes the MIMIC dataset to evaluate model performance across diverse clinical contexts, thereby demonstrating the framework's broad generalization capabilities. These findings support the viability of deploying such pre-trained models in general medical practice, although the slightly elevated calibration error highlights an area for future refinement. The pre-training dataset consisted of patients of Asian origin, whereas the validation datasets were drawn from populations in the United States (osteoporosis and MIMIC datasets) and Australia (thyroid disease dataset). These datasets differed not only geographically but also in demographic composition, providing context to assess the generalization and transferability of GDP across diverse diseases and populations. This phenomenon is analogous to the cross-lingual capabilities of language models \cite{pires2019multilingual,nllb2024scaling,siddhant2020xtreme}, which adapt to new languages with minimal or no target-language supervision. Related transfer phenomena have also been observed in healthcare: for instance, the cross-disease transfer of laboratory trajectories \cite{MyGLP}, as well as cross-modality transfer in tasks such as restoring low-quality ECG signals \cite{liu2024ecg} and decoding neural signals to interpret brain activity with language models \cite{lee2024enhancing}. Collectively, these properties suggest that medical-specific foundation models hold considerable promise for addressing data scarcity and imbalance across populations and diseases.

When discussing the constitutional configuration of GDP, our findings confirm that deep learning methods are highly dependent on spatial information, as variations in input ordering can produce substantially different outcomes \cite{borisov2022deep,ghosh2019investigating}. The encoding methods, on the other hand, demonstrates inconsistent results, generally, $trad$-based encoding demonstrates better and more stable performance. 

The empirical success of this work validates our design objectives by demonstrating that GDP-derived embeddings provide superior representational capacity, thereby enhancing demographic feature contributions. By generating enriched representations, our pre-trained model successfully amplifies the predictive utility of demographic variables beyond their raw form. Crucially, these embeddings enhance prediction outcomes across diverse disease settings and patient populations. Furthermore, the proposed methodology yields complementary representations that can be directly concatenated with residual features, operating as a plug-and-play enhancement layer compatible with standard tabular classifiers. Grounded in these empirical results, future work will aim to optimize model calibration and discrimination under conditions of high patient variability, while extending pre-trained models to additional EHR modalities to accelerate the development and clinical adoption of medical AI systems.

\section{Conclusion}

This study explores the development of a pre-trained model tailored specifically to demographic attributes, with the expectation that such a model can provide generalizable enhancements to predictive performance across diverse medical tasks and geographic population. Aiming to strike a balance between generality and task specificity, this work proposes the segmentation of EHR components to generate complementary embedding representations. These embeddings can be directly concatenated with residual features, operating as a plug-and-play enhancement layer that remains fully compatible with standard tabular classifiers. Our proposed methodology elevates the feature importance of demographic attributes and yields competitive performance that surpasses top-performing tabular foundation models. Furthermore, the resulting embeddings exhibit transferability, generalized across varying clinical environments and patient cohorts. Ultimately, this framework offers a highly plug-and-play solution to enhance existing healthcare models, maintaining computational simplicity and seamless compatibility with established machine learning classifiers.

\section*{Acknowledge}
This work was supported by the National Science and Technology Council, Taiwan, NSTC 113-2410-H-002-273-. The funder played no role in study design, data collection, analysis and interpretation of data, or the writing of this manuscript. All authors declare no financial or non-financial competing interests. Gemini 3.5 flash was used for correcting grammar and English improving in the manuscript.

\section*{Appendix}
\subsection*{Configuration for LightGBM}
LightGBM was configured for classification, with 50 estimators, the 'gbdt' boosting type, and a learning rate of 0.1. For t-SNE, the embedded space dimensionality was set to 2, with a perplexity value of 5. The hyperparameter settings were held constant across all experiments. MIMIC dataset constitute of patient data extracted from the MIMIC-III database \cite{mimiciii}. Neonates, patients older than 90, and individuals with incomplete temporal records (missing birth, measurement, admission, or discharge dates) were excluded from the study. The final feature set includes demographic attributes (age and sex) and six key laboratory variables: Chol/HDL-c ratio, LDL-c level, LDL-c/HDL-c ratio, fasting blood glucose, WBC count, and UA level. The primary prediction target was defined as patient death within 30 days of discharge. For the pneumonia dataset, multiple records per patient were aggregated into a single observation using median values, and missing data were imputed using an Iterative Imputer with a Random Forest estimator\footnote{\url{https://scikit-learn.org/stable/modules/generated/sklearn.impute.IterativeImputer.html}}. 

\subsection*{Age-sex interaction analysis}
\begin{figure}
\centerline{\includegraphics[width=0.8\columnwidth]{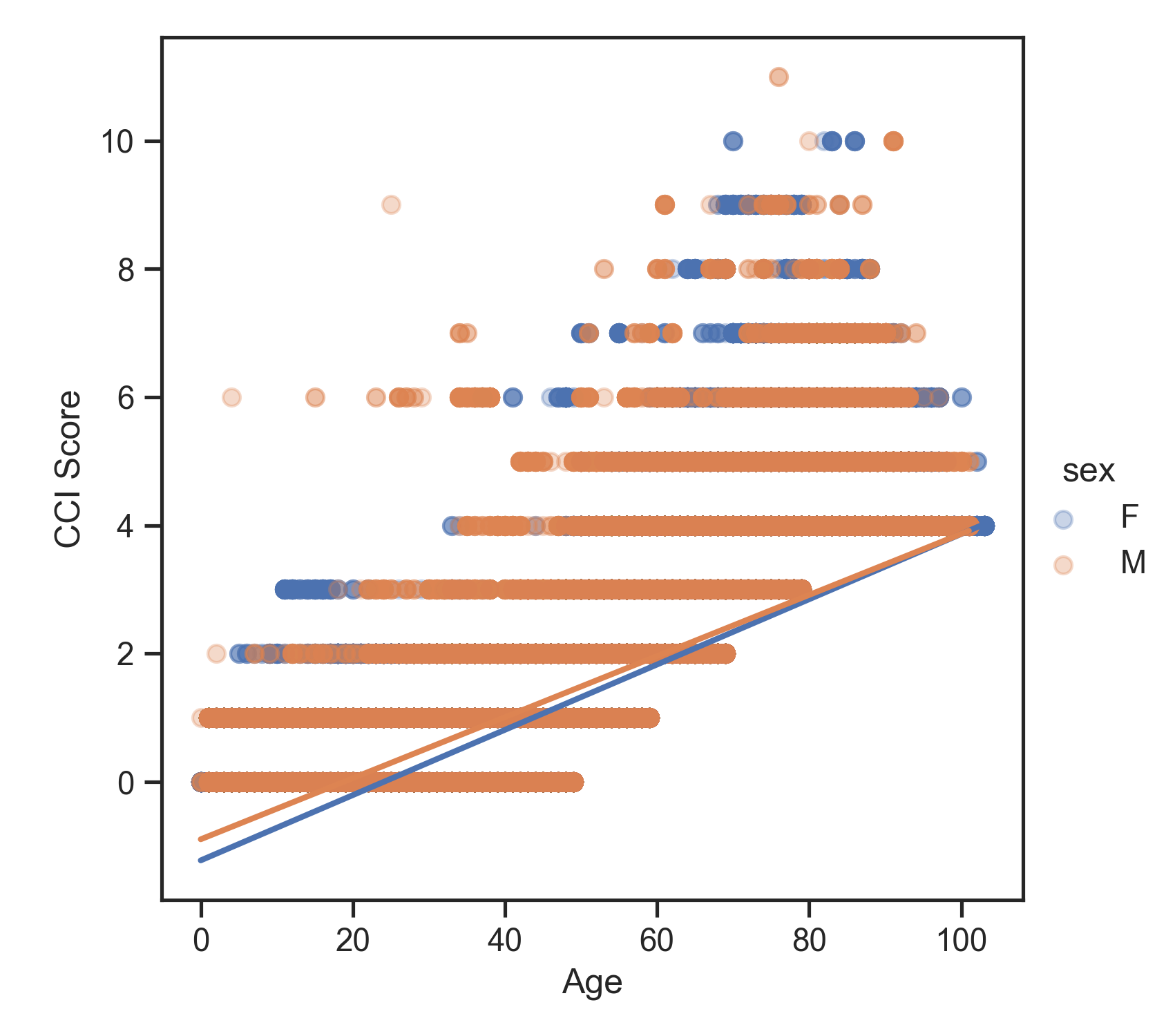}}
\caption{Linear Age-Sex Interaction}
\label{age_linear}
\end{figure}

\begin{figure}
\centerline{\includegraphics[width=0.8\columnwidth]{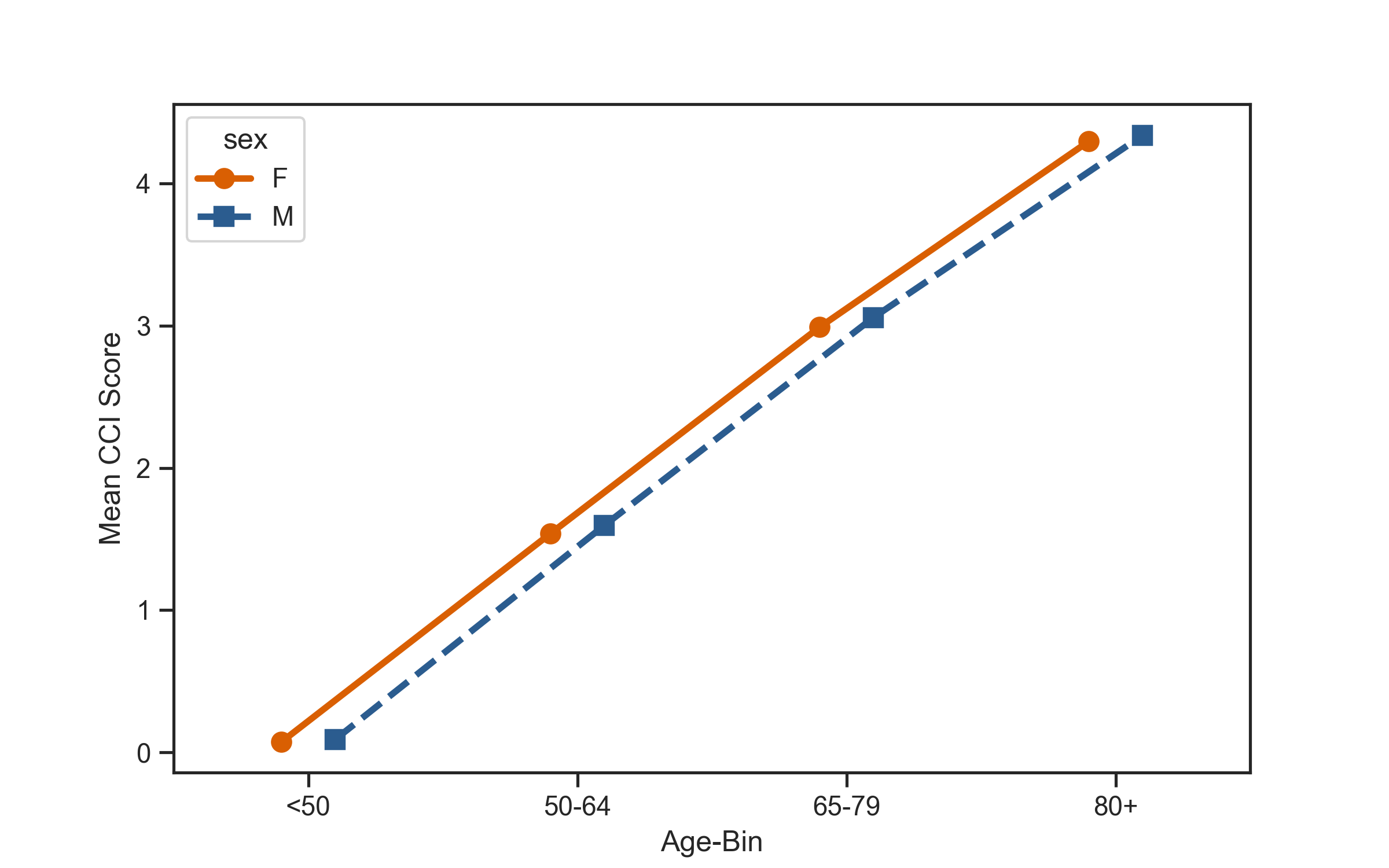}}
\caption{Age-Bin x Sex Interaction}
\label{age_bin}
\end{figure}

\begin{figure}
\centerline{\includegraphics[width=0.8\columnwidth]{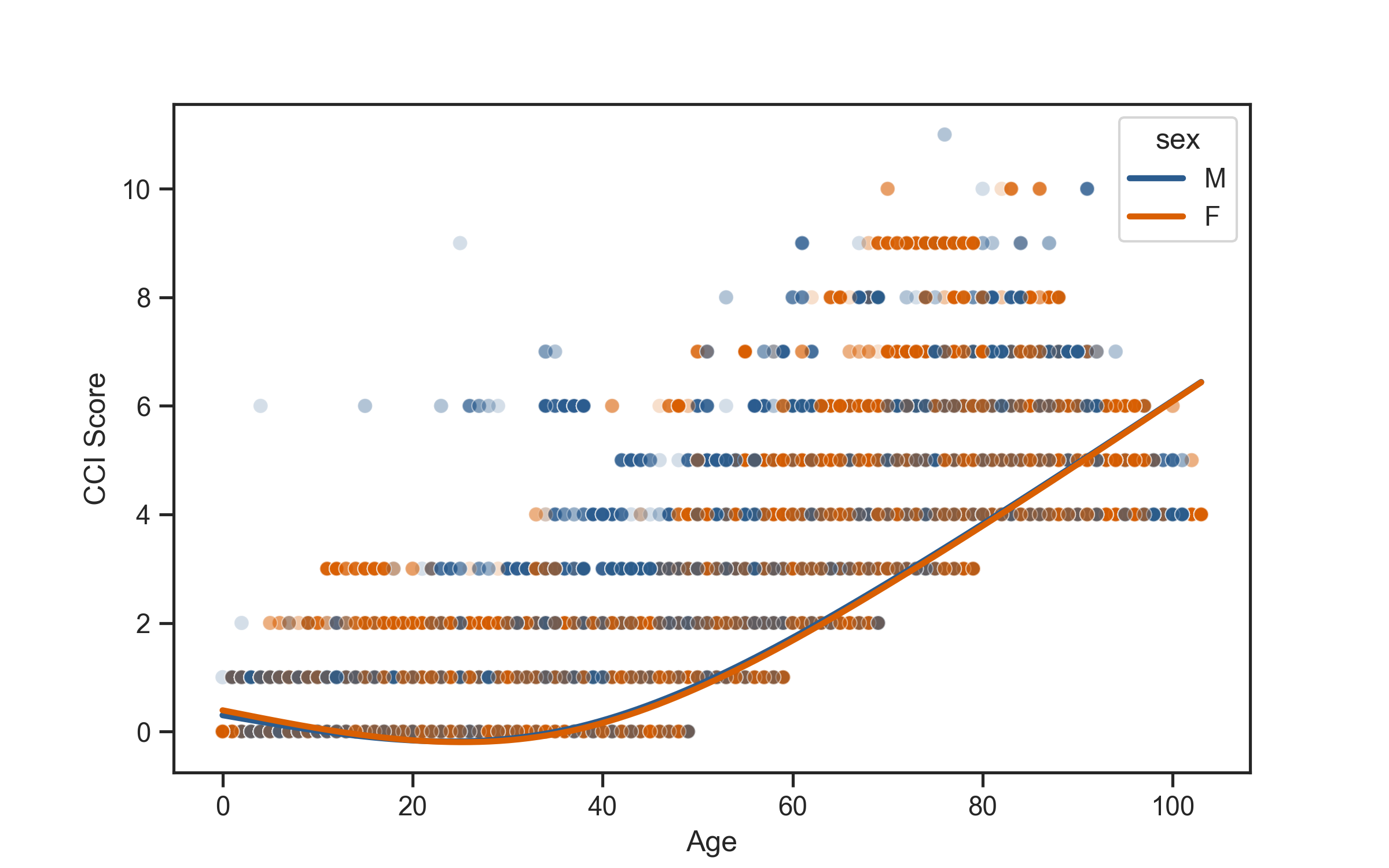}}
\caption{Spline Age-Sex Interaction}
\label{age_spline}
\end{figure}

In the multivariable model $\textit{CCI} \sim \textit{age} + \textit{sex}$, age was the primary determinant of CCI, with each additional year of age associated with a 0.049-point increase in CCI score ($\beta = $ 0.049, p $<$ 0.001). Independent of age, male sex was associated with a statistically significant but clinically minor elevation in baseline CCI score ($\beta = $0.190, p $<$ 0.001). As visualized across linear (Figure~\ref{age_linear}), binned (Figure~\ref{age_bin}), and spline (Figure~\ref{age_spline}) trajectories, aging influences comorbidity accumulation at an identical rate for both sexes, with males maintaining a small, parallel elevation of 0.190 points across the lifespan.

\bibliographystyle{plain}
\bibliography{references}


\end{document}